\documentclass[letterpaper, 10 pt, conference]{ieeeconf}  

\IEEEoverridecommandlockouts                              

\usepackage{graphics} 
\usepackage{epsfig} 
\usepackage{times} 
\usepackage{amsmath} 
\usepackage{amssymb}  
\usepackage{color}
\usepackage{amsfonts}

\usepackage{graphicx}
\def\BibTeX{{\rm B\kern-.05em{\sc i\kern-.025em b}\kern-.08em
    T\kern-.1667em\lower.7ex\hbox{E}\kern-.125emX}}
\usepackage[font=small,labelfont=bf,tableposition=top]{caption}

\usepackage{blindtext}

\title{\LARGE \bf
Accurate Instance-Level CAD Model Retrieval\\in a Large-Scale Database
}

\author{Jiaxin Wei$^{1}$, Lan Hu$^{1,2}$, Chenyu Wang$^{1}$, Laurent Kneip$^{1}$
\thanks{$^{1}$Mobile Perception Lab, School of Information Science and Technology, ShanghaiTech University; \newline {\tt\small \{weijx,wangchy4,lkneip\}@shanghaitech.edu.cn}}
\thanks{$^{2}$Shanghai Institute of Microsystem and  Information Technology, Chinese Academy of Sciences and University of Chinese Academy of Sciences. {\tt\small hulan@shanghaitech.edu.cn}}
}

\let\oldtwocolumn\twocolumn
\renewcommand\twocolumn[1][]{%
    \oldtwocolumn[{#1}{
    \begin{center}
        \includegraphics[width=0.99\textwidth]{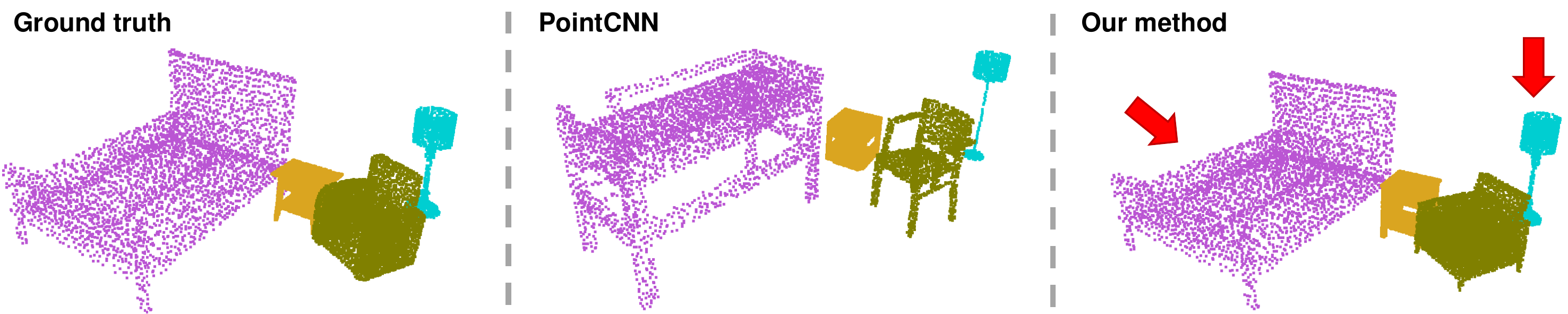}
        \captionof{figure}{Scene reconstruction of scene0143\_00 from the Scan2CAD \cite{Scan2CAD} dataset. This scene contains a bed, a table, a chair, and a lamp. Compared to PointCNN \cite{pointcnn}, our method not only retrieves the ground truth model for the bed and the lamp, but also gives more reasonable results for the table and the chair.}
        \label{Fig:scene}
    \end{center}
    }]
}

\begin{document}

\maketitle
\thispagestyle{empty}
\pagestyle{empty}

\begin{abstract}

We present a new solution to the fine-grained retrieval of clean CAD models from a large-scale database in order to recover detailed object shape geometries for RGBD scans. Unlike previous work simply indexing into a moderately small database using an object shape descriptor and accepting the top retrieval result, we argue that in the case of a large-scale database a more accurate model may be found within a neighborhood of the descriptor. More importantly, we propose that the distinctiveness deficiency of shape descriptors at the instance level can be compensated by a geometry-based re-ranking of its neighborhood. Our approach first leverages the discriminative power of learned representations to distinguish between different categories of models and then uses a novel robust point set distance metric to re-rank the CAD neighborhood, enabling fine-grained retrieval in a large shape database. Evaluation on a real-world dataset shows that our geometry-based re-ranking is a conceptually simple but highly effective method that can lead to a significant improvement in retrieval accuracy compared to the state-of-the-art.

\end{abstract}

\section{INTRODUCTION}

Nowadays, emerging robotics and augmented reality~\cite{VRAR} applications have a huge demand for real-time 3D reconstructions of object-level scene models from RGBD scans. Traditional incremental fusion solutions~\cite{kinectfusion}\cite{fusion++} often fail to recover detailed scene geometries due to the noisy nature of the measurements captured by consumer-grade hardware. Instead of focusing on the exhaustive refinement of 3D scans \cite{revealnet}, an increasing number of approaches \cite{alignmodel}\cite{Scan2CAD}\cite{end2end} turn to an offline CAD database for better reconstructions. Given sufficient object detection, segmentation, and representation abilities, the original incomplete scans may be greatly improved by substituting corresponding measurement parts with similar CAD models. 

However, the imperfection of object detection and segmentation, together with the incapability of 3D shape descriptors to encode precise geometry information, jointly place a major restriction on CAD model retrieval at a finer level. Current methods, such as \cite{Scan2CAD}\cite{end2end}, concentrate on computing the 9DoF alignment between scanned objects and CAD models while only requiring the retrieval of a same-category model due to the usage of a small ground truth pool. Zhao \textit{et al.} \cite{CORSAIR} and Dahnert \textit{et al.} \cite{JointEmbed} claim retrieval of CAD instances with higher similarity, but they still constrain the model search to a relatively small database containing an augmented ground truth set for each scanned object. We instead focus on instance-level retrieval of CAD models in a large shape database, which is more practical in real-world scenarios.

CAD models from the same category usually possess a universal geometric structure. Therefore, it is quite easy for learned representations \cite{pointnet}\cite{pointnet2}\cite{dgcnn} to achieve outstanding performance in 3D shape classification. Yet, individual CAD models within the same class vary drastically in terms of shape details, making it hard to differentiate them from one another. To this end, we propose a two-step pipeline to tackle this challenging problem. We first retrieve multiple CAD candidates from the shape descriptor space using nearest neighbor search, and then use geometric residuals to re-rank those candidates, thereby obtaining much better results both quantitatively and qualitatively. Our method is based on two important insights. First, the discriminative power of learned features can help narrow down the search to models with the same semantic class as that of the scanned object. And a suitable hierarchical data structure on top of our large-scale database makes the search more efficiently. Second, after figuring out which part of a database is of interest, we rely on point set distance metrics to further perform a local search and refine the ranking order of initial candidates. We also suggest a new variant of Chamfer Distance to deal with pathological cases in evaluating candidates.

The main contributions of this paper are summarized as follows:
\begin{itemize}
    \item We present a new pipeline for instance-level CAD model retrieval at large scale. It relies on a feature-based $k$-nearest neighbor search followed by a geometry-based re-ranking, which can considerably improve the quality of retrieved CAD models.
    \item We introduce a variant of the Single-direction Chamfer Distance \cite{CORSAIR} that is more robust to outliers when measuring point set distances between partial scans and CAD models.
\end{itemize}

As demonstrated by our experiments, the combination of geometric re-ranking outperforms the plain application of state-of-the-art feature embedding networks. From a qualitative perspective, it also produces much more visually similar results, and thus for the first time enables instance-level retrieval of CAD models within a large-scale database.

\section{Related Work}
\textbf{CAD Model Retrieval:}
There are mainly two directions for 3D model retrieval: view-based methods and model-based methods. View-based methods extract and merge 2D features from multiple projections of a 3D object and merge those features to generate a global shape descriptor for later recognition while model-based methods directly deal with 3D data. View-based methods are out of scope of this paper. The interested readers are referred to \cite{grabner2019location}\cite{bansal2016marr}\cite{he2018triplet} for more details.

Most traditional 3D model retrieval algorithms \cite{kim2012acquiring}\cite{li2015database}\cite{kim2013guided}\cite{drost20123d} are based on hand-designed descriptors \cite{fpfh}\cite{johnson1997spin} that may easily suffer from the inconsistency between clean CAD models and noisy real-world data. Meanwhile, learning-based methods are getting more and more popular. Scan2CAD \cite{Scan2CAD} carries out implicit retrieval by evaluating the semantic compatibility between CAD models and a patch from the scan. Avetisyan \textit{et al.}~\cite{end2end} insert an extra object detection module such that the relevant parts of a scan can be directly used to calculate descriptors. Dahnert \textit{et al.} \cite{JointEmbed} segment the foreground from the input scan and complete the segmented objects by stacking hourglass encoder-decoders, resulting in a joint embedding of 3D scans and CAD models. More recently, CORSAIR~\cite{CORSAIR} extends FCGF~\cite{fcgf} to simultaneously learn a global feature for retrieval as well as local point features for alignment. 

The aforementioned methods have shown promising results yet are not practical. They either assume a small pool of exact CAD models as input or consider a pool with augmented ground truth models for each scanned object. In contrast, our proposed method performs retrieval in a more realistic setting where the database has thousands of models and a one-to-many relationship with scanned objects.

\textbf{3D Shape Descriptors:}
Given that there are plenty of excellent works for instance segmentation \cite{chen2020blendmask}\cite{bolya2019yolact}\cite{he2017mask}, we can safely assume that objects have already been segmented from scans. 3D model retrieval can thus be achieved by learning a global shape descriptor for object scans, which has attracted increasing attention over the past few years. For example, VoxNet \cite{voxnet} applies a supervised 3D CNN on volumetric occupancy grids to predict object class labels. Alternative approaches directly work on point clouds. PointNet \cite{pointnet} extracts point features using shared MLPs and aggregates them with a max pooling operator to obtain a permutation-invariant global feature. Later on, Qi \textit{et al.} propose PointNet++ \cite{pointnet2} to further capture the local geometric structures of point clouds from neighborhood points. In the spirit of graph neural networks, DGCNN \cite{dgcnn} constructs a local graph on the point cloud which can be dynamically updated using aggregated edge features after each layer of the network. As a generalization of 2D CNN, PointCNN \cite{pointcnn} uses $\mathcal{X}$-Conv to transform the input point clouds and raw features into a canonical form and then applies typical convolutions. However, those works usually do not work well with noisy or partial measurements. We resort to geometric information to reduce the uncertainty.

\textbf{Set Similarity for Point Clouds:}
Fan \textit{et al.} \cite{PointGen} propose two distance metrics to compare generated point clouds against ground truth: Chamfer Distance (CD) and Earth Mover's Distance (EMD). EMD gives the minimum effort of transforming one point set into the other and has been proven to be more accurate than CD in point cloud generation \cite{PointGen}\cite{LearningRG}. The biggest issue impeaching EMD from widespread use is the high computational cost for exact solutions. CD is preferred \cite{Deng2018PPFFoldNetUL}\cite{Deng20193DLF}\cite{Duan20193DPC} when efficiency is of major concern.

\section{Method}

\begin{figure*}[htbp]
    \centering
    \includegraphics[scale=0.45]{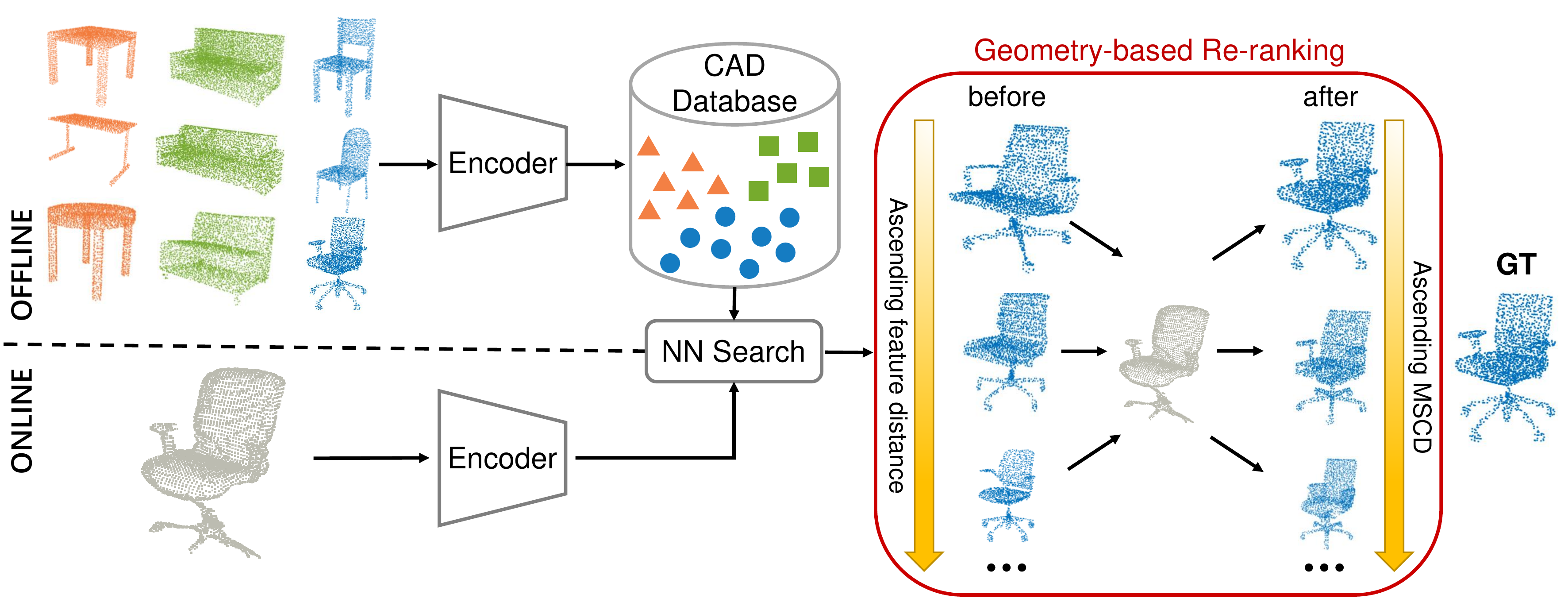}
    \caption{An illustration of our pipeline. CAD models are processed offline to form a large database in which we can do $k$-nearest neighbor search for each query object. Based on the coarsely retrieved CAD candidates, we further perform geometry-based re-ranking using Modified Single-Direction Chamfer distance (MSCD) to make sure that higher-ranked candidates are generally more similar to the query object.}
    \label{Fig:pipeline}
\end{figure*}

This section is organized as follows. Section \ref{Sec:overview} reveals our motivation of establishing instance-level retrieval in a large CAD database and provides an overview of our method. In Section \ref{Sec:rank} and \ref{Sec:rerank}, we describe our two-step pipeline in all details, comprising of feature-based ranking and geometry-based re-ranking.

\subsection{Overview}\label{Sec:overview}
The retrieval of a similar CAD model is tightly entangled with the cross-instance alignment between a CAD model and a scanned object.
With the rise of category-level object alignment \cite{FCGFalignment}, it is possible to align different object instances within a category. For ease of implementation, we simply assume that all the retrieved CAD models are already aligned with the query object by leveraging such alignment methods.

Avetisyan \textit{et al.} \cite{Scan2CAD}\cite{end2end} retrieve CAD models from a small pool tailored for each scene. The model pool contains only the ground truth CAD models corresponding to the objects that are effectively present in the scene. On the contrary, we perform retrieval in a large-scale CAD database, which is more closer to the real situation. We furthermore focus on instance-level rather than category-level retrieval. For category-level retrieval, the matching of categories between the retrieved CAD and the query object is enough to be regarded as a success. However, real-world applications such as robot interactions with surrounding environment require a detailed understanding of object shapes. In other words, the retrieved CAD model should replicate the object shape as closely as possible. In contrast to the works of Ishimtsev \textit{et al.} \cite{CADeform} and Uy \text{et al.} \cite{JointDeform}, our method avoids the involvement of expensive CAD model deformation, and makes full use of a large-scale database to identify the model that is most similar to the scanned object.

Our method breaks down into two procedures. First, we calculate a shape descriptor using a neural network and then perform coarse retrieval by means of $k$-nearest neighbour search to obtain an initial set of CAD candidates. It points out regions of interest in the database that are highly likely to contain the best shape. Second, we apply geometric verification to assess and re-rank the initial candidates. As it is hard to evaluate the similarity between retrieved CAD models and incomplete scanned objects, we introduce a modified version of Chamfer Distance to better handle the outliers. This step can significantly eliminate the uncertainty incurred by neural networks. In addition, the refinement is quite efficient given that the number of CAD candidates is far fewer that of models in the whole database. Our complete pipeline is shown in Figure \ref{Fig:pipeline}.

\subsection{Feature-Based Ranking}\label{Sec:rank}
Although neural networks may not yet be able to capture the exact geometry of an object, it is still easy for them to categorize shapes into the right semantic classes. Consequently, we utilize the discriminative power of learned global features to characterize inter-class dissimilarity. All the query objects and CAD models are processed through the same backbone encoder. Note that there are little constraints on the network architecture as long as it outputs a global feature for each shape. Here we adopt neural networks originally designed for 3D shape classification.

Given a query object $\mathbf{X} \in \mathbb{R}^{N \times 3}$ and a CAD database $\mathcal{Y}=\{\mathbf{Y}_i | \mathbf{Y}_i \in \mathbb{R}^{M_i \times 3}, 1 \leq i \leq S\}$ consisting of $S$ models, all represented in the form of point clouds, our goal is to retrieve a CAD model $\mathbf{Y} \in \mathcal{Y}$ that is not only semantically compatible to $\mathbf{X}$ but also geometrically close to it. Let $f(\cdot)$ denote the backbone encoder. It takes as input a point cloud $\mathbf{P}$ and outputs a $d$-dimensional shape descriptor
\begin{equation}
f: \mathbf{P} \mapsto f(\mathbf{P}) \in \mathbb{R}^d \notag.
\end{equation}
Out of consideration for efficiency, features for all CAD models in the database are pre-computed offline. Let $\mathbf{F}^\mathbf{X} \in \mathbb{R}^d$ and $\mathbf{F}^\mathcal{Y}=\{\mathbf{F}^\mathcal{Y}_i | \mathbf{F}^\mathcal{Y}_i \in \mathbb{R}^d, 1 \leq i \leq S\}$ be the extracted global features for the query object and the models, respectively. Each time a new query arrives, we perform a $k$-nearest neighbor search using the object feature to obtain an ordered candidate set:
\begin{equation}
\mathcal{C}=\{\mathbf{C}_j | \mathbf{C}_j \in \mathcal{Y}, 1 \leq j \leq k\} \notag
\end{equation}
such that $d(\mathbf{F}^\mathbf{X}, \mathbf{F}^\mathcal{C}_j)$ is the $j$-th smallest element in $d(\mathbf{F}^\mathbf{X}, \mathbf{F}^\mathcal{Y})$. Here $d(\cdot)$ represents the Euclidean distance between feature vectors. To reduce the search complexity, we explore the database via hierarchical data structures, such as a kd-tree.

\subsection{Geometry-Based Re-ranking}\label{Sec:rerank}
In most cases, the initial set may contain a very similar model but its ranking order is unsatisfactory owing to the limited ability of learned features to encode precise shape information. The distinctiveness can be further brought down when some parts of the query object are occluded or unobserved. Meanwhile, the bounded number of CAD models in the database means that there usually do not exist identical matches. Therefore, we exploit geometric information to address the intra-class similarity. Specifically, we re-rank the initial CAD candidates by measuring the geometric difference between point clouds. 

Chamfer Distance (CD) is a popular point set distance metric used in many learning based point cloud generation frameworks \cite{PointGen}\cite{LearningRG}. For two point clouds $P$ and $Q$, a common definition of CD is to find the nearest neighbor of $P$ in $Q$ and vice versa, and then add up all squared distances. That is,
\begin{equation}
d_{CD}(P, Q) = \frac{1}{|P|}\sum_{p \in P}\min_{q \in Q}\|p-q\|_2^2 + \frac{1}{|Q|}\sum_{q \in Q}\min_{p \in P}\|p-q\|_2^2,
\label{Eq:two_way_chamfer}
\end{equation}
where $|P|$ and $|Q|$ are the number of points in $P$ and $Q$, respectively. The division by $|P|$ and $|Q|$ acts as a normalization which can balance the influence caused by different point densities in $P$ and $Q$. Note that the term \textit{Chamfer Distance} is a slight abuse of notation as it does not actually satisfy the triangle inequality required by a proper distance metric~\cite{PointGen}.

An interesting variant of CD has been introduced in \cite{CORSAIR}, called the Single-direction Chamfer Distance (SCD). It simply drops one side of squared distances in (\ref{Eq:two_way_chamfer}), resulting in
\begin{equation}
d_{SCD}(P, Q) = \frac{1}{|P|}\sum_{p \in P}\min_{q \in Q}\|p-q\|_2^2. \label{Eq:one_way_chamfer}
\end{equation}
Hence, SCD only maintains the distances from scanned object to CAD model. It is intuitively clear that this simplification helps SCD to concentrate on the visible parts of a scanned object while neglecting unobserved parts.

For a better understanding of (\ref{Eq:one_way_chamfer}), let us define a distance vector
\begin{equation}
  \mathbf{d} = \left[\begin{matrix}\min_{q \in Q}\|p_1-q\|_2 \\ \min_{q \in Q}\|p_2-q\|_2 \\ ... \\
  \min_{q \in Q}\|p_{|P|}-q\|_2 \end{matrix} \right] \in \mathbb{R}^{|P|}. \notag
\end{equation}
Thus, SCD can be reformulated as 
\begin{equation}
    d_{SCD}(P, Q) = \frac{1}{|P|}\|\mathbf{d}\|_2^2, \label{Eq:one_way_chamfer_reform}
\end{equation}
the $L_2$-norm of the distance vector $\mathbf{d}$. However, it is well known from the literature of robust optimization that the $L_2$-norm is sensitive to outliers as it emphasises on residuals by squaring them. In practical scenarios, object scans often contain outlier measurements owing to the existence of segmentation errors and other artifacts such as floating point measurements. In order to improve the robustness of the above distance against such artifacts, we propose to use the $L_1$-norm of $\mathbf{d}$ instead. We denote this distance the Modified Single-direction Chamfer Distance (MSCD), and it is given by
\begin{equation}
d_{MSCD}(P, Q) = \frac{1}{|P|}\sum_{p \in P}\min_{q \in Q}\|p-q\|_2. \label{Eq:one_way_chamfer_no_square}
\end{equation}
Though the change seems subtle, our experimental results demonstrate that MSCD achieves better results in practice.

To conclude, we take each candidate CAD model from $\mathcal{C}$ and calculate the MSCD with respect to our object scan, yielding a new ordered set
\begin{equation}
\hat{\mathcal{C}}=\{\hat{\mathbf{C}}_j | \hat{\mathbf{C}}_j \in \mathcal{C}, 1 \leq j \leq k\} \notag
\end{equation}
such that $d_{MSCD}(\mathbf{X}, \hat{\mathbf{C}}_j)$ is the $j$-th smallest element in $d_{MSCD}(\mathbf{X}, \mathcal{C})$.

\section{Experiments}

\begin{table*}[htbp]
\centering
\caption{Top1/Top5 retrieval accuracy}
\label{Tab:accuracy}
\begin{tabular}{|l|cccccccc|cc|}
\hline
Method         & bed       & bookshelf & cabinet   & chair     & display   & sofa      & table     & other     & \textbf{class avg.} & \textbf{instance avg.} \\ \hline\hline
VFH \cite{vfh}            & 0.00/0.00 & 0.00/0.02 & 0.00/0.00 & 0.01/0.02 & 0.01/0.03 & 0.00/0.01 & 0.01/0.04 & 0.00/0.00 & 0.00/0.01  & 0.01/0.02     \\
DGCNN \cite{dgcnn}          & 0.04/0.17 & 0.02/0.10 & 0.00/0.06 & 0.01/0.04 & 0.01/0.04 & 0.01/0.07 & 0.01/0.04 & 0.06/0.24 & 0.25/0.47  & 0.02/0.09     \\
VoxNet \cite{voxnet}         & 0.12/0.37 & 0.06/0.17 & 0.02/0.06 & 0.02/0.05 & 0.03/0.08 & 0.04/0.12 & 0.03/0.10 & 0.07/0.22 & 0.22/0.50  & 0.04/0.11     \\
PointNet2 \cite{pointnet2}      & 0.10/0.28 & 0.02/0.08 & 0.01/0.05 & 0.01/0.05 & 0.01/0.07 & 0.02/0.08 & 0.02/0.08 & 0.09/0.28 & 0.29/0.51  & 0.03/0.11     \\
PointCNN \cite{pointcnn}       & 0.05/0.29 & 0.16/0.31 & 0.04/0.11 & 0.02/0.08 & 0.04/0.11 & 0.02/0.10 & 0.05/0.15 & 0.11/0.32 & 0.29/0.57  & 0.06/0.16     \\ \hline
VFH+MSCD       & 0.01/0.01 & 0.05/0.14 & 0.02/0.03 & 0.02/0.08 & 0.03/0.07 & 0.03/0.05 & 0.08/0.11 & 0.01/0.01 & 0.02/0.04  & 0.03/0.07     \\
DGCNN+MSCD     & \textbf{0.42}/0.74 & 0.28/0.53 & 0.18/0.31 & 0.08/0.17 & 0.10/0.26 & 0.18/0.33 & 0.13/0.21 & \textbf{0.30/0.63} & 0.39/0.70  & 0.17/0.33     \\
VoxNet+MSCD    & 0.41/\textbf{0.75} & 0.26/0.51 & 0.21/0.43 & 0.08/0.18 & 0.11/0.29 & \textbf{0.25}/0.40 & 0.17/0.31 & \textbf{0.30}/0.62 & \textbf{0.43}/0.69  & 0.18/0.36     \\
PointNet2+MSCD & 0.41/0.72 & 0.28/0.53 & 0.23/0.44 & 0.10/0.21 & 0.12/0.29 & 0.22/0.37 & 0.19/0.32 & \textbf{0.30/0.63} & \textbf{0.43}/0.70  & 0.19/0.37     \\
PointCNN+MSCD  & 0.41/0.72 & \textbf{0.29/0.56} & \textbf{0.27/0.54} & \textbf{0.11/0.26} & \textbf{0.16/0.35} & \textbf{0.25/0.42} & \textbf{0.27/0.47} & \textbf{0.30/0.63} & \textbf{0.43/0.71}  & \textbf{0.22/0.43}     \\ \hline
\end{tabular}
\end{table*}

This section is organized as follows. Section \ref{Sec:dataset} introduces the dataset on which we conduct all our experiments. Section \ref{Sec:comparisons} describes the alternative state-of-the-art methods and Section \ref{Sec:metrics} defines three evaluation metrics. Discussions of the performance and ablation studies are presented in Section \ref{Sec:results} and Section \ref{Sec:ablation}, respectively. Finally, in Section \ref{Sec:qualitative}, we show some qualitative resutls to further demonstrate the superiority of our method.

\subsection{Dataset}\label{Sec:dataset}
We test our method on the real-world dataset Scan2CAD~\cite{Scan2CAD}, which contains 14225 (3049 unique) CAD models from ShapeNet \cite{shapenet} and corresponding scanned objects from ScanNet \cite{scannet}. For CAD model retrieval, we assume that instance segmentations are known, and apply the segmentation masks over the 3D scans to obtain object surfaces. We furthermore sample points on the surface of the CAD models to generate point clouds for the database. Note that we filter out some poor quality point clouds with too few points. Finally, we work with 12598 object scans spreading over 1506 scenes, and a database of 2827 CAD models.

\subsection{Shape Descriptor Alternatives}\label{Sec:comparisons}
For comparison purposes, we choose several state-of-the-art 3D shape descriptors. This includes the handcrafted feature VFH~\cite{vfh} as well as the learning-based features DGCNN~\cite{dgcnn}, VoxNet~\cite{voxnet}, PointNet2 (or PointNet++)~\cite{pointnet2}, and PointCNN~\cite{pointcnn}. Viewpoint Feature Histogram (VFH) is a global descriptor based on FPFH~\cite{fpfh}. We compute the 308-dimensional VFH feature using the PCL implementation \cite{pcl}. As for DGCNN, VoxNet, PointNet2 and PointCNN, they are representatives of graph-based, volumetric-based, point-based, and convolution-based methods for 3D shape classification, respectively. All networks are trained on ModelNet40 \cite{modelnet40} and we extract the global feature from the input of the final classification layer. Both DGCNN and PointNet2 output 256-dimensional feature vectors while VoxNet and PointCNN produce 128-dimensional features.

\subsection{Evaluation Metrics}\label{Sec:metrics}
We use three different metrics to evaluate the retrieval performance on Scan2CAD:

\noindent\textbf{Topk Retrieval Accuracy.} Suppose we have $T$ query objects in $\mathcal{X}$ and there is a ground truth CAD model $\mathbf{Y}^* \in \mathcal{Y}$ for each query object $\mathbf{X} \in \mathcal{X}$. The Topk retrieval accuracy (RA) is defined as
\begin{equation}
\mathrm{RA} =  \frac{1}{T}\sum_{\mathbf{X} \in \mathcal{X}} \mathbf{1}_{\mathcal{A}_k}(\mathbf{Y}^*), \label{Eq:RA}
\end{equation}
where $\mathcal{A}_k$ can be replaced either by the top-k subset of the initial ordered set $\mathcal{C}$, or the top-k subset of the re-ordered set $\hat{\mathcal{C}}$. Note that $\mathbf{1}_{\mathcal{A}_k}: \mathcal{Y} \rightarrow \{0, 1\}$ is an indicator function:
\begin{equation}
\mathbf{1}_{\mathcal{A}_k}(\mathbf{Y}^*) = 
\left\{
\begin{aligned}
1 \quad & \mathrm{if~} \mathbf{Y}^* \in \mathcal{A}_k, \\
0 \quad & \mathrm{otherwise}. 
\end{aligned}
\right. \notag
\end{equation}

\noindent\textbf{Top1 Chamfer Distance.} With a slight modification of the metric defined in \cite{CORSAIR}, the Top1 Chamfer Distance is computed by measuring MSCD between the ground truth CAD model and the retrieved top 1 CAD model:
\begin{equation}
d = \frac{1}{T}\sum_{\mathbf{X} \in \mathcal{X}} d_{MSCD}(\mathcal{A}_1, \mathbf{Y}^*). \label{Eq:d}
\end{equation}

\noindent\textbf{Ground truth Ranking.} The ranking order of the ground truth CAD model in the returned CAD candidate set:
\begin{equation}
\mathrm{R} = \frac{1}{T}\sum_{\mathbf{X} \in \mathcal{X}} \mathrm{ranking}_{\mathcal{A}}(\mathbf{Y}^*) \label{Eq:ranking}
\end{equation}
where $\mathrm{ranking}_{\mathcal{A}}(\cdot)$ denotes the ranking position in $\mathcal{A}$.

\subsection{Results}\label{Sec:results}
We use the methods mentioned in Section \ref{Sec:comparisons} as our front-end, followed by MSCD-based re-ranking to improve the CAD candidate set returned by the initial feature-based search. Note that the nearest neighbour search is configured to find 90 nearest neighbors. Table \ref{Tab:accuracy} indicates Top1 and Top5 retrieval accuracy of different methods and demonstrates the impact of adding MSCD-based re-ranking. It is obvious that the proposed geometric re-ranking strategy leads to a significant performance boost over the original feature-based ranking results. In addition, our method achieves much better results when combined with PointCNN.

\begin{table}[b]
\centering
\caption{Top5 Category Ratio}
\label{Tab:category}
\begin{tabular}{|c|c|}
\hline
Method    & Top5 Category Ratio \\ \hline\hline
VFH \cite{vfh}       & 0.32                \\
DGCNN \cite{dgcnn}     & 0.99                \\
VoxNet \cite{voxnet}    & \textbf{1.00}       \\
PointNet2 \cite{pointnet2} & 0.99                \\
PointCNN \cite{pointcnn}  & 0.99                \\ \hline
\end{tabular}
\end{table}

\begin{figure}[htbp]
    \centering
    \includegraphics[width=0.35\textwidth]{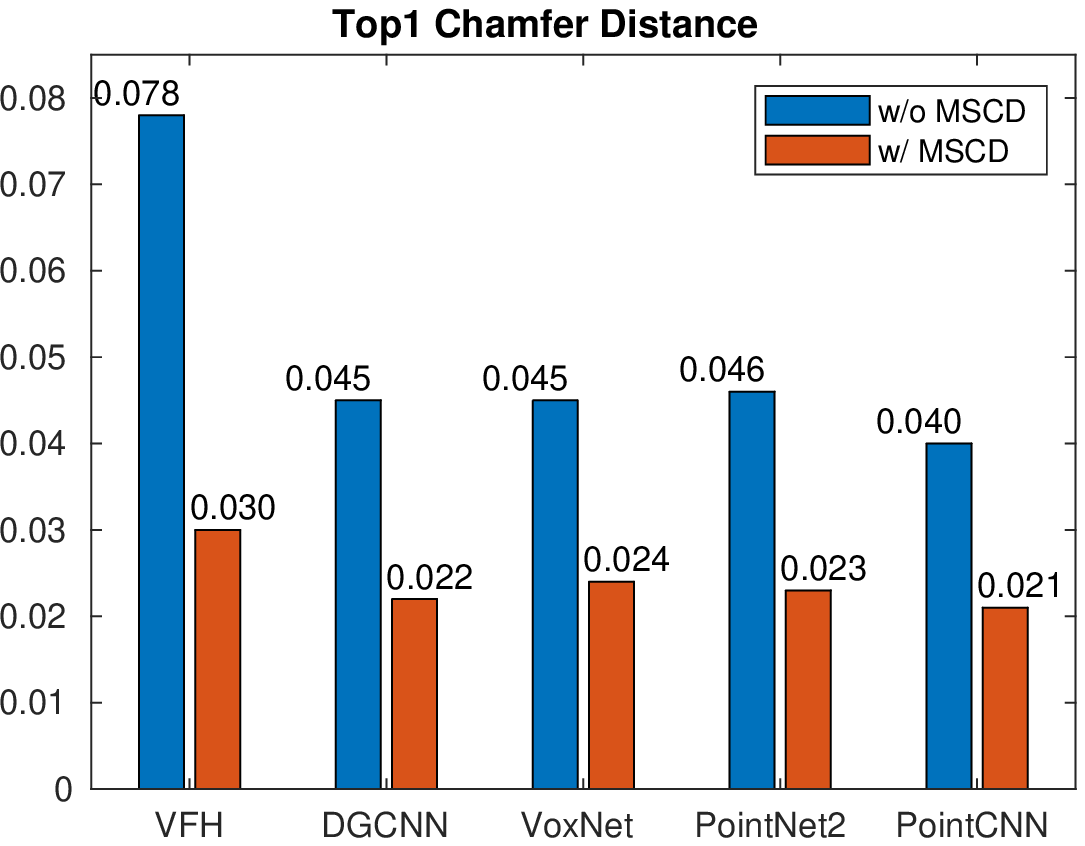}
    \caption{Comparative results of Top1 Chamfer Distance on Scan2CAD.}
    \label{Fig:chamfer}
\end{figure}

\begin{figure}[htbp]
    \centering
    \includegraphics[width=0.35\textwidth]{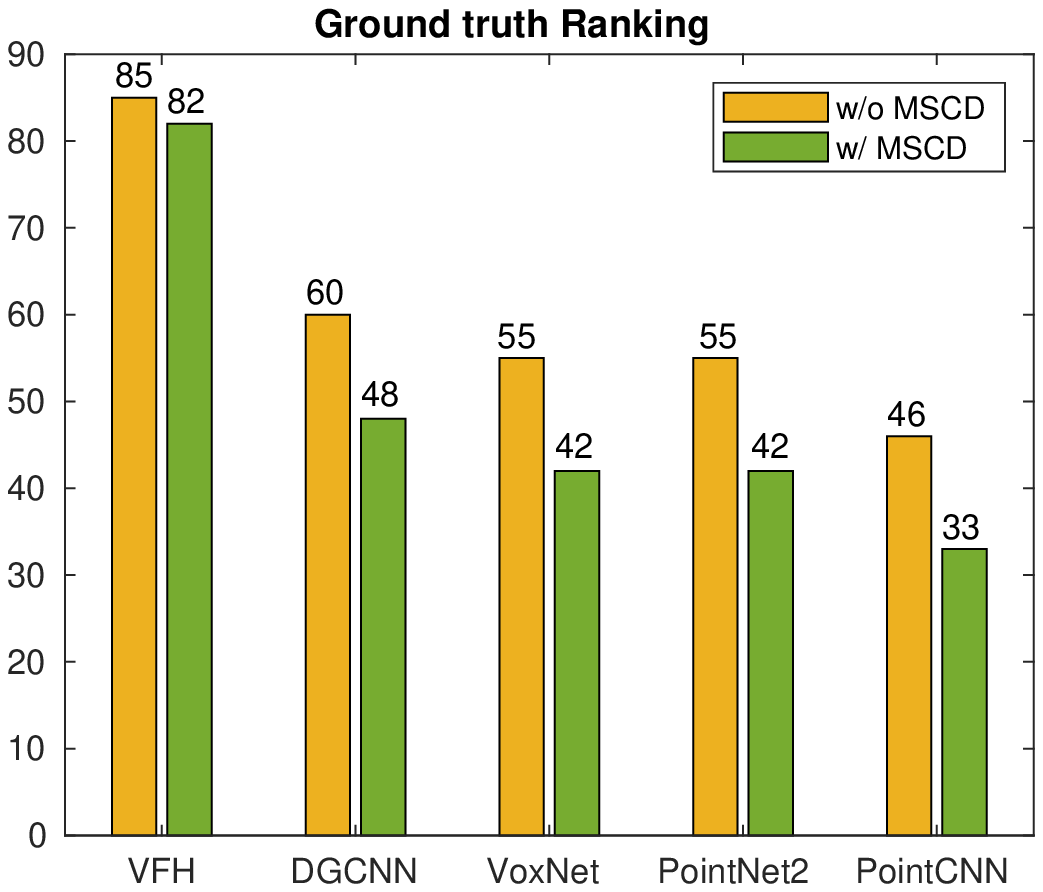}
    \caption{Comparative results of Ground truth Ranking on Scan2CAD.}
    \label{Fig:ranking}
\end{figure}

\subsection{Ablation Studies}\label{Sec:ablation}

\begin{figure}[htbp]
    \centering
    \includegraphics[width=0.4\textwidth]{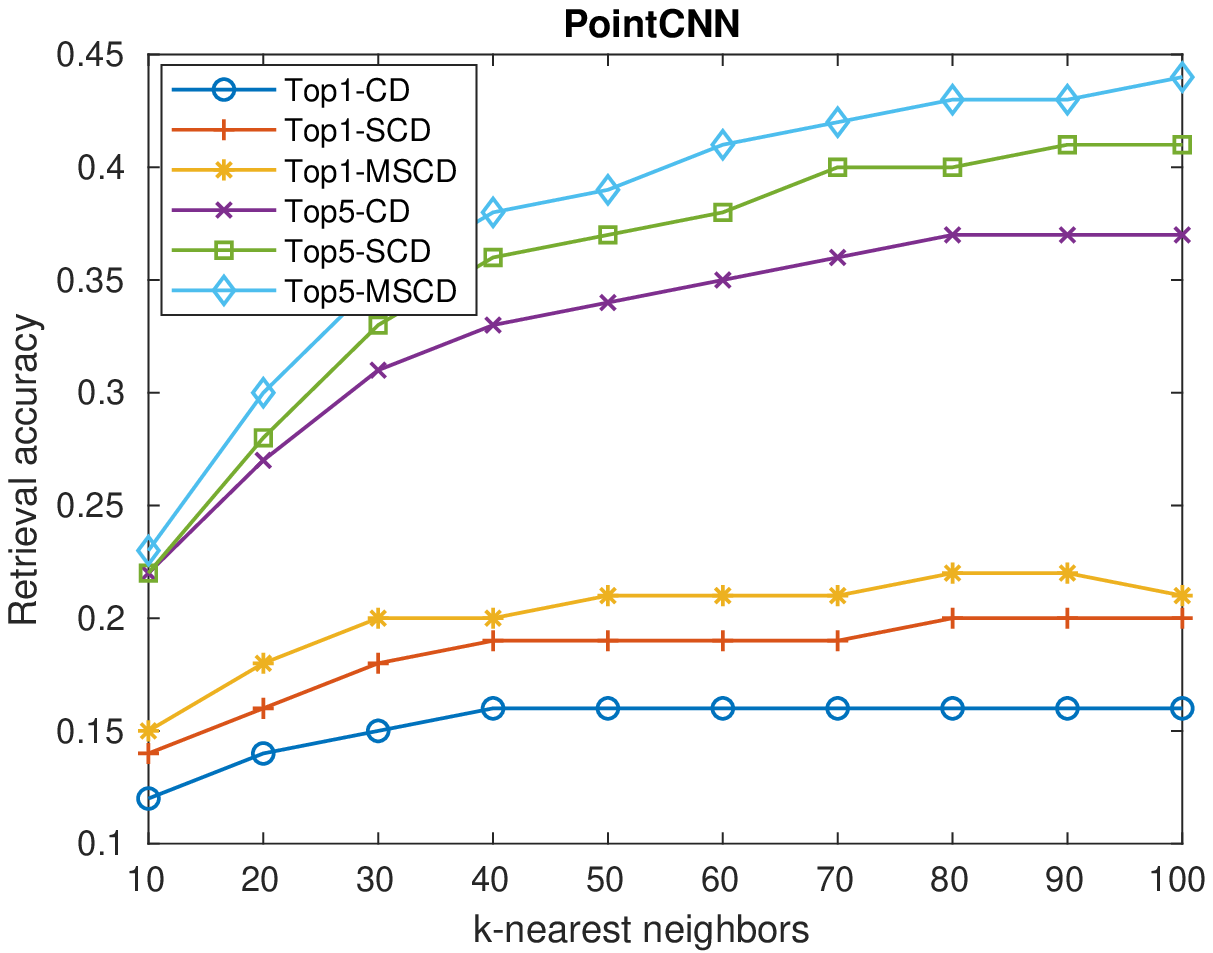}
    \caption{Top1 and Top5 retrieval accuracy using different metrics and different search ranges based on PointCNN.}
    \label{Fig:metrics}
\end{figure}

\begin{figure*}[htbp]
    \centering
    \includegraphics[scale=0.14]{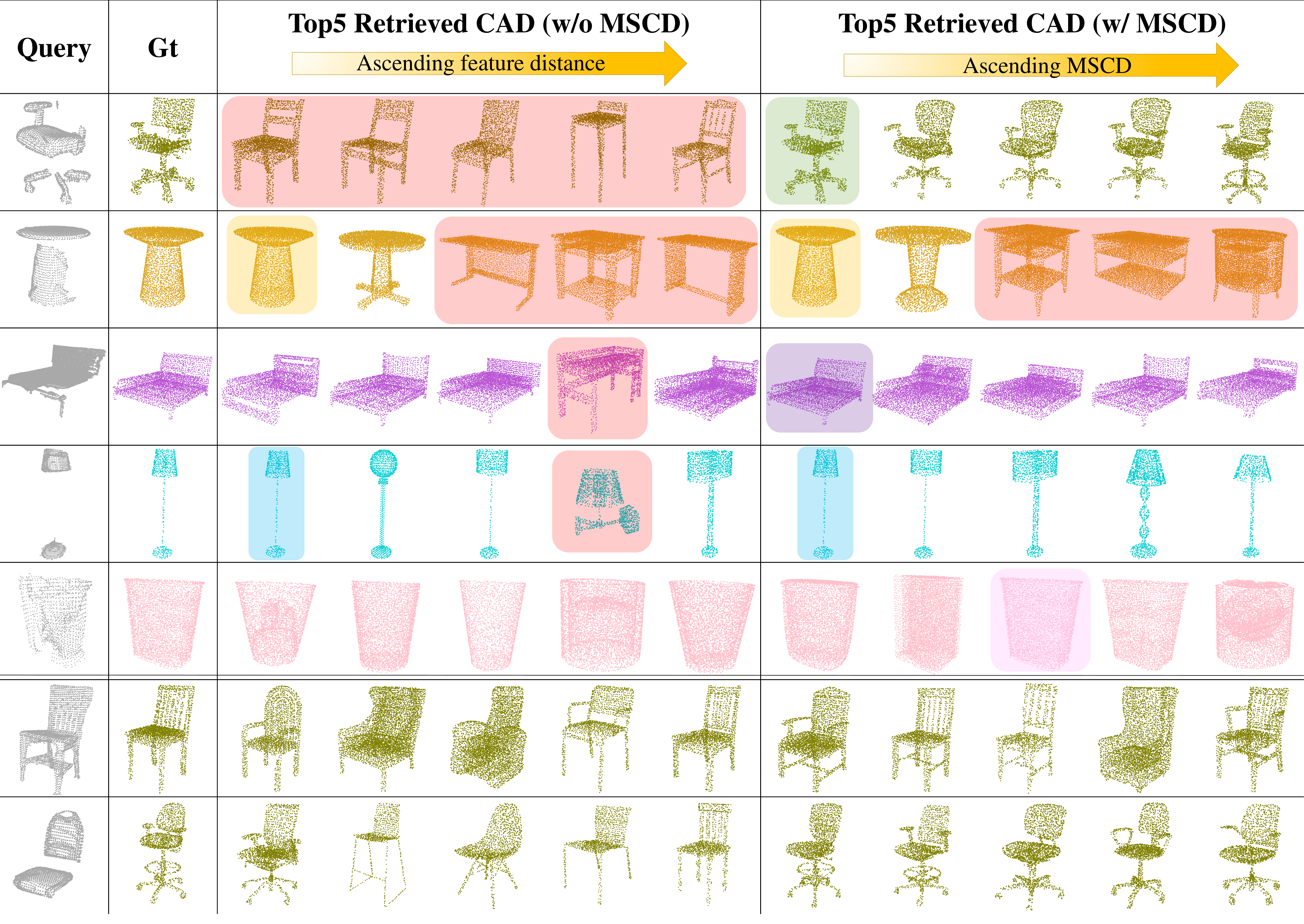}
    \caption{Qualitative results of CAD model retrieval on Scan2CAD \cite{Scan2CAD} based on PointCNN \cite{pointcnn}. Rows 1 to 5 show different kinds of examples for classes chair, table, bed, lamp and trash bin, respectively. The retrieved CAD model identical to ground truth is highlighted with corresponding color while the models of clearly inferior quality are highlighted with red. The last two rows are failure cases.}
    \label{Fig:qualitative}
\end{figure*}

In Table \ref{Tab:category}, we show that the top 5 CAD models produced by 3D shape descriptors almost always have the same category as that of the query object, only with the exception of VFH. This proves the usefulness of neural networks in handling inter-class dissimilarity and also explains the poor performance of VFH.

We further test our MSCD-based re-ranking method using Top1 Chamfer Distance. As shown in Figure \ref{Fig:chamfer}, the MSCD between the ground truth and the retrieved top 1 CAD model declines for all 3D shape descriptors when combining with a re-ranking method. We also compare the rank of the ground truth CAD model with and without re-ranking to validate its effectiveness (see Figure \ref{Fig:ranking}). After re-ranking, the ground truth CAD model is promoted to a higher rank.

Our method has very few hyper-parameters. To figure out the influence of hyper-parameters on retrieval performance, we run a series of experiments based on PointCNN using different nearest neighbor search ranges and different point cloud distance metrics (introduced in Section \ref{Sec:rerank}). Figure \ref{Fig:metrics} shows that MSCD outperforms both CD and SCD in terms of Top1 and Top5 retrieval accuracy. The performance gradually improves as the search range becomes larger.

\subsection{Qualitative Results}\label{Sec:qualitative}
Qualitative results for CAD model retrieval on Scan2CAD are shown in Figure \ref{Fig:qualitative}. Our method takes good advantage of visible parts in the query object and retrieves more reasonable CAD models with similar structure. For example, let us consider the first row in Figure \ref{Fig:qualitative}. The wheel and armrest fragments in the query indicate that this is a swivel chair with armrests. It is obvious that our proposed re-ranking method returns CAD models with both of these properties while the feature-based search struggles to get reasonable results. Furthermore, our method successfully rejects models of clearly inferior quality (cf. row 3 and row 4 of Fig. \ref{Fig:qualitative}).

The last two rows in Figure \ref{Fig:qualitative} show failure cases of our approach. Most of them are caused by the inherent ambiguity in query objects and inaccurate ground truth annotations. Taking the bottom row as an example, the query barely captures the back and the seat of a chair, which creates possibilities for arbitrary types of legs. Though our method retrieves some similar CAD models, it fails to find the exact model of the ground truth.

\section{Conclusion}
We have presented a two-step pipeline for instance-level CAD model retrieval in a large-scale database. Since state-of-the-art learned representations still have limitations on understanding all aspects of 3D shapes, we have proposed a geometry-based re-ranking method to facilitate fine-grained retrieval. We have furthermore introduced a Modified Single-direction Chamfer Distance to measure the point set distance between partial objects and CAD models. Thorough experiments on real-world test cases validate the superiority of our method. Future work will investigate the combination of instance-level retrieval and category-level alignment in order to simultaneously perform CAD model retrieval and alignment with high accuracy.


\bibliographystyle{IEEEtran}
\bibliography{ref}

\end{document}